\documentclass{article}

\usepackage[preprint]{corl_2023} 

\usepackage{times}

\usepackage[numbers]{natbib}
\usepackage{multicol}

\usepackage{graphicx}
\usepackage{amsmath}
\usepackage{amssymb}
\usepackage{booktabs}
\usepackage{algorithm}
\usepackage{algpseudocode}
\usepackage{xspace}
\usepackage{amsthm}
\usepackage{tablefootnote}

\usepackage{amsmath,amsfonts,bm}

\def\eqref#1{equation~\ref{#1}}

\def\1{\bm{1}}

\def\vmu{{\bm{\mu}}}

\def\va{{\bm{a}}}

\def\ve{{\bm{e}}}

\def\vh{{\bm{h}}}

\def\vp{{\bm{p}}}

\def\vs{{\bm{s}}}

\def\vv{{\bm{v}}}

\def\mD{{\bm{D}}}

\def\mI{{\bm{I}}}

\def\mW{{\bm{W}}}

\DeclareMathAlphabet{\mathsfit}{\encodingdefault}{\sfdefault}{m}{sl}
\SetMathAlphabet{\mathsfit}{bold}{\encodingdefault}{\sfdefault}{bx}{n}

\usepackage{textcomp}
\usepackage{siunitx}
\usepackage{authblk}
\usepackage[T1]{fontenc}
\usepackage{babel}
\newcommand*{\affaddr}[1]{#1} 
\newcommand*{\affmark}[1][*]{\textsuperscript{#1}}
\newcommand*{\email}[1]{\texttt{#1}}

\makeatletter
\DeclareRobustCommand\onedot{\futurelet\@let@token\@onedot}
\def\@onedot{\ifx\@let@token.\else.\null\fi\xspace}

\makeatother

\usepackage[capitalize]{cleveref}
\crefname{section}{Sec.}{Secs.}
\Crefname{section}{Section}{Sections}
\Crefname{table}{Table}{Tables}
\crefname{table}{Tab.}{Tabs.}
\crefname{algorithm}{Algo.}{Algos.}
\crefname{equation}{Eq.}{Eq.}

\title{Long-term Microscopic Traffic Simulation with History-Masked Multi-agent Imitation Learning}

\author{%
Ke Guo\affmark[1,2], Wei Jing\affmark[2], Lingping Gao\affmark[2], Weiwei Liu\affmark[2], and Weizi Li\affmark[3], Jia Pan\affmark[1]\\
\affaddr{\affmark[1] The University of Hong Kong}, \email{\{kguo, jpan\}@cs.hku.hk}\\
\affaddr{\affmark[2] Alibaba Group  \email{\{21wjing,gaolingping.glp\}@gmail.com, 11932061@zju.edu.cn}} \\
\affaddr{\affmark[3] University of Memphis  \email{wli@memphis.edu}} \\
}

\begin{document}
\maketitle


\begin{abstract}
A realistic long-term microscopic traffic simulator is necessary for understanding how microscopic changes affect traffic patterns at a larger scale. Traditional simulators that model human driving behavior with heuristic rules often fail to achieve accurate simulations due to real-world traffic complexity. To overcome this challenge, researchers have turned to neural networks, which are trained through imitation learning from human driver demonstrations. However, existing learning-based microscopic simulators often fail to generate stable long-term simulations due to the \textit{covariate shift} issue. To address this, we propose a history-masked multi-agent imitation learning method that removes all vehicles' historical trajectory information and applies perturbation to their current positions during learning. We apply our approach specifically to the urban traffic simulation problem and evaluate it on the real-world large-scale pNEUMA dataset, achieving better short-term microscopic and long-term macroscopic similarity to real-world data than state-of-the-art baselines.
\end{abstract}


\section{Introduction}

Microscopic traffic simulators are powerful tools for transportation engineers and planners to analyze and predict the impact of microscopic adjustments on traffic patterns without disrupting real-world traffic. For example, it can help analyze how changing road shape like replacing an intersection with a roundabout affects traffic patterns~\cite{Bared2005SimulatedCO}, and develop traffic-aware autonomous driving policies that enhance overall traffic efficiency~\cite{zheng2023trafficdriving,Wang2023LearningTC}. However, creating a realistic simulator that can simultaneously replicate the microscopic response of human drivers to traffic conditions and the resulting long-term macroscopic statistics is a challenging task.

In recent years, there have been significant efforts to develop realistic traffic simulators that accurately model human driving behavior. Traditional traffic simulators, such as SUMO~\citep{krajzewicz2012sumo}, AIMSUN~\citep{barcelo2005AIMSUN}, and MITSIM~\citep{yang1996microscopic}, typically rely on heuristic car-following models like the Intelligent Driver Model (IDM)~\citep{treiber2000idm}. However, despite careful calibration of parameters, these simplified, rule-based models often fail to deliver accurate simulations~\citep{Feng2021IntelligentDI} due to the complexity of real-world traffic environments. Factors such as road structure, neighboring vehicles, and even driver psychology can influence human driver decision-making, making it challenging to achieve accurate simulations.

To improve the capabilities of traffic simulators, researchers have turned to neural networks as a driving model. These models are trained through imitation learning (IL) from human driver demonstrations. Most studies~\citep{morton2016analysis,Suo2021TrafficSim} employ behavior cloning (BC)~\citep{Michie1990BC} to learn the driving policy in a supervised fashion by minimizing the difference between the model output and the human driver's action in the training state distribution. However, the BC method suffers from the \textit{covariate shift} issue~\citep{ross2011reduction}, where the state induced by the learner's policy deviates cumulatively from the expert's distribution.

To address this issue, recent works~\citep{song2018multi} propose using generative adversarial imitation learning (GAIL)~\citep{ho2016generative}. GAIL learns a reward function by a discriminator neural network and trains the policy network to maximize the reward through online reinforcement learning. This allows agents to learn to recover from out-of-distribution states. However, directly applying GAIL to the traffic simulation problem, which is a multi-agent imitation learning task, is problematic because the environment changes during the policy learning process, leading to highly biased estimated gradients. To mitigate this instability, the parameter-sharing generative adversarial imitation learning (PS-GAIL)~\citep{song2018multi} makes all agents share the policy and critic parameter and gradually increases the agent numbers. However, even in a simple highway environment, there are many undesirable off-road driving cases~\citep{Bhattacharyya2019SimulatingEP}, which limit the effectiveness of PS-GAIL.

Although existing learning-based microscopic simulators have shown success in short-term simulation applications, such as autonomous driving tests~\citep{Suo2021TrafficSim,Bergamini2021SimNetLR}, they often fail to generate stable long-term traffic simulations. Hence, we propose a history-masked multi-agent imitation learning (HMMIL) method that can remove all agents' historical trajectory information and apply a perturbation to their current positions during learning. Our method is inspired by context-conditioned imitation learning (CCIL)~\cite{Guo2023CCIL}, a single-agent offline imitation learning method for autonomous driving. 

CCIL solves the \textit{covariate shift} issue in BC for autonomous driving by removing the ego vehicle's historical trajectories and adding perturbations to its current position because the ego state is highly susceptible to policy errors, while human drivers in the context are assumed to be robust to the ego vehicle's policy error. However, directly applying CCIL to the multi-agent imitation learning problem by making all agents share the same policy works poorly because other vehicles' behaviors are also directly determined by the learned policy, unlike in the autonomous driving task. This leads to an additional \textit{covariate shift} in the context. To overcome this problem, our method removes the histories of all vehicles and adds perturbation to all vehicles instead of only the ego vehicle in CCIL. 

The main contributions of our paper are:
\begin{itemize}
    \item We propose a new history-masked multi-agent imitation learning (HMMIL) method that can address the \textit{covariate shift} issue in multi-agent imitation learning.
    \item We apply our approach specifically to urban traffic simulation. To the best of our knowledge, this is the first imitation learning-based traffic simulator that can reproduce long-term (more than 10 minutes) microscopic urban traffic.
    \item We evaluate our method on a real-world large-scale dataset, named pNEUMA~\citep{barmpounakis2020pNEUMA}, achieving better short-term microscopic and long-term macroscopic similarity to real-world data than state-of-the-art baselines. 
\end{itemize}

\section{Related Work}

While macroscopic traffic simulators in~\citep{Lighthill1955OnKW,Richards1956ShockWO,sewall2010continuum} can efficiently reproduce long-term macroscopic statistics, they are incapable of analyzing the effects of microscopic changes. Therefore, we focus on microscopic traffic simulators that can capture detailed interactions between vehicles and accurately replicate human driving behavior. These simulators can be classified into rule-based and learning-based simulators based on their driving models.

\subsection{Rule-based Simulator}

Rule-based models like the IDM~\citep{treiber2000idm} and Krauss model~\citep{krauss1998microscopic} have been widely applied in popular traffic simulators such as SUMO~\citep{krajzewicz2012sumo}, AIMSUN~\citep{barcelo2005AIMSUN}, and MITSIM~\cite{yang1996microscopic}. These models describe individual vehicle behavior based on a car-following model that predicts the longitudinal acceleration of a vehicle based on its relative speed and distance to its front vehicle. Although there are some parameter calibration methods~\citep{kesting2008calibrating,osorio2019efficient} using real data, these models are oversimplified in their assumptions of interactions between traffic participants and are therefore limited in their accuracy.

\subsection{Learning-based Simulator}

To improve the modeling capacity and similarity to human behavior, recent studies have attempted to learn a neural driving model by imitation from human driving demonstrations. These methods can be generally classified into BC-based and GAIL-based methods.

BC-based methods, such as TrafficSim~\citep{Suo2021TrafficSim} and SimNet~\citep{Bergamini2021SimNetLR}, usually first learn a prediction model and then modify the predicted trajectories to avoid collisions and traffic rule violations during simulation. However, these BC-based methods cannot achieve long-term simulation due to the \textit{covariate shift} problem caused by the discrepancy between the distribution of the training data and the learned policy's state distribution. In contrast, our method is learned offline like BC, but we address the \textit{covariate shift} problem by ignoring historical trajectories and blurring current positions, thus achieving long-term stable simulation. To improve performance, we also modify the predicted trajectory by projecting it onto the road and making it smooth from the current state during simulation, but skip the computationally costly collision removal operation because we mainly focus on the long-term macroscopic influence.

GAIL-based methods~\citep{song2018multi,Bhattacharyya2019SimulatingEP,Koeberle2022ExploringTT} learn the hidden reward function of human driving behavior and obtain the driving policy by maximizing the learned reward. While GAIL can theoretically address the \textit{covariate shift} of BC in a single-agent context by online interaction, its performance deteriorates when applied to the multi-agent imitation learning domain due to the dynamic environment, leading to a tricky training process. To address this issue, PS-GAIL~\citep{song2018multi} requires two-stage learning and gradually adds vehicles to the environment. However, PS-GAIL still exhibits a significant number of undesirable traffic phenomena, such as off-road driving, collisions, and hard braking. Based on PS-GAIL, the reward-augmented imitation learning (RAIL) method~\citep{Bhattacharyya2019SimulatingEP,Koeberle2022ExploringTT} penalizes undesirable phenomena by adding a hand-crafted reward, but maximizing the new reward does not guarantee the recovery of human-like trajectories. Despite many improvements on the original GAIL, these GAIL-based methods usually fail to produce stable long-term traffic flow, as demonstrated in our experiments. In contrast, our method is an offline supervised learning method, leading to faster, simpler, and more stable learning of human driving policy.

\section{Method}

\begin{figure*}[t]
	\centering
	\includegraphics[width=\linewidth]{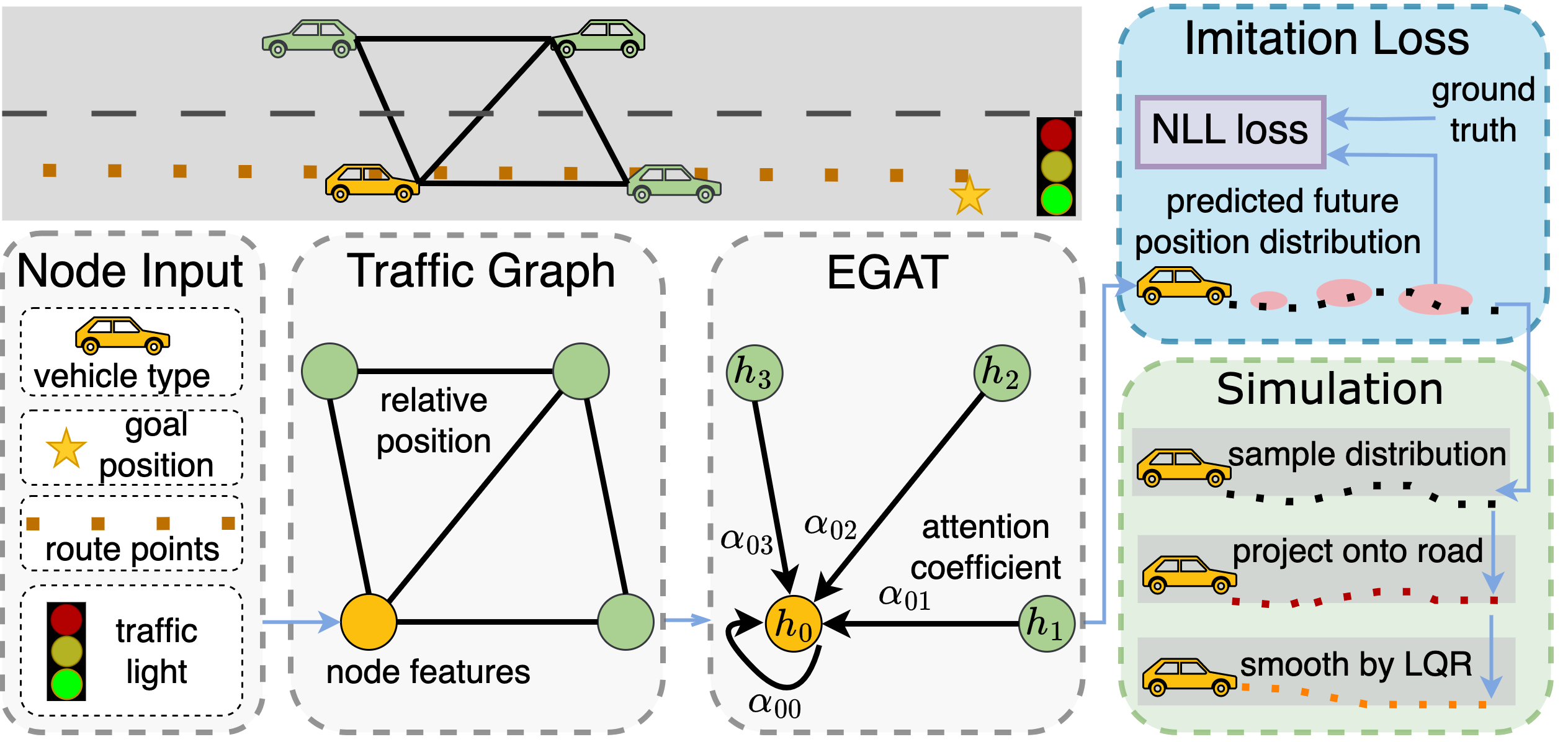}
	\vspace{-0.7cm}
	\caption{Overview of our approach.}
	\vspace{-0.5cm}
	\label{fig:overview}
\end{figure*}

In~\cref{fig:overview}, we present an overview of our method. Our method predicts vehicles' future trajectory distribution using a graph neural network, where their history information is overlooked during learning. During simulation, we yield a smooth action for each vehicle based on sampled positions from the predicted distribution and its current state. 

\subsection{Traffic Graph Representation}

In the traffic system, a human driver makes decisions mainly depending on its neighboring context including other vehicles and road network. Therefore, we can model the system with a graph by connecting neighboring elements. 

\textbf{Node:} To save computational cost, we only create agent nodes and assign road information to them. Each agent node has input features including its type, destination position, nearest routing points with corresponding road width, and the traffic light status of its closest road. The agent's history is not taken into account to avoid the \textit{covariate shift} issue. Because each agent's routing information contains partial road network information, each agent can still obtain its neighboring road network information by exchanging information with neighboring agents. To improve the model's generalizability, we transform each node feature to its individual coordinate system. We use the ego-perturbed goal-oriented coordinate system described in CCIL~\citep{Guo2023CCIL}, where the origin is the agent's current position plus a zero-mean Gaussian perturbation, and the $x$-axis direction points towards its goal position.

\textbf{Edge:} When transforming each agent's state coordinates from the global coordinate system to their individual coordinate system, information about the relative positions among agents is lost. However, a traffic model needs to consider the relative configuration of agents to understand how they interact with each other. To preserve these relationships, we introduce directed edges among neighboring agents. The edge feature is the relative position of the destination node in the source node's coordinate system. In practice, we connect an agent with its nearest 8 neighbors located within a neighborhood distance of 30 meters.

\subsection{Edge-enhanced Graph Attention Network}

To predict the future position distribution of all agents, we utilize an edge-enhanced graph attention network (EGAT)~\citep{diehl2019graph,Mo2022MultiAgentTP}. EGAT is a variant of graph attention networks (GAT)~\citep{velivckovicgraph} that can efficiently model interactions by aggregating neighboring node information using an attention mechanism. However, traditional GAT can only handle node features, while our traffic graph's edges contain relative position information. EGAT overcomes this challenge by concatenating edge features with connected node features to perform aggregation. In each EGAT layer, the node's next state is calculated as follows:
\begin{equation}
    \vh_i^l=\sigma \left(\sum_{j \in \mathcal{N}_i} \alpha_{i j}^l \mW^l \left[\vh^{(l-1)}_i \|  \ve_{i j}\|\vh^{(l-1)}_j \right]\right),
\end{equation}
where $\vh_i^l$ is the feature of node $i$ in the $l$ th layer, $\ve_{ij}$ is the position of node $j$ relative to node $i$, $\mW^l$ is the learnable weight matrix, $\mathcal{N}_i$ is the set of the first-order neighbors of node $i$ (including the node itself), and $\sigma$ is a non-linear activation function. The node features at the first layer are obtained by embedding the node input features with multi-layer perceptions (MLPs). The attention coefficient $\alpha_{ij}^l$ indicates the importance of node $j$ to node $i$, considering both node and edge features, and is computed as:
\begin{equation}
\alpha_{i j}^l=\operatorname{softmax}_j \left(\sigma\left((\va^l)^T\left[\vh^{(l-1)}_i \|  \ve_{i j}\|\vh^{(l-1)}_j \right]\right)\right),
\end{equation}
where $\va^l$ is a learnable weight vector, and normalization is performed on the weights across all neighbors of node $i$ using a $\operatorname{softmax}$ function.

\subsection{Imitation Loss}
After passing through multiple EGAT layers, the hidden state of each node $m$ is fed into a fully connected layer to predict its future position distribution over $T$ time steps, denoted as $p(\hat{\vp}_1^m,\hat{\vp}_2^m,...,\hat{\vp}_T^m)$, which is assumed to be a product of multi-variable Gaussian distributions:
\begin{equation}
p(\hat{\vp}_1^m,\hat{\vp}_2^m,...,\hat{\vp}_T^m)=\prod_{t=1}^{T} \mathcal{N}(\hat{\vmu}^m_t, \hat{\boldsymbol{\Sigma}}^m_t),
\end{equation}
where $\hat{\vmu}^m_t$ and $\hat{\boldsymbol{\Sigma}}^m_t$ represent the mean and covariance matrix of the predicted position $\hat{\vp}_t^m$ at future time step $t$, respectively. For simplicity, we assume that there is no correlation between the position distributions at different future time steps. To learn the graph neural network, we minimize the negative log-likelihood (NLL) loss of all agents' ground-truth future trajectories:
\begin{equation}
    \mathcal{L}=-\sum_{m=1}^{M}\sum_{t=1}^{T} \log(\mathcal{N}(\vp_t^{m}-\hat{\vmu}^m_t, \hat{\boldsymbol{\Sigma}}^m_t)) ,
\end{equation}
where $\vp_t^{m}$ denotes the ground truth position of agent $m$ at future time step $t$, and $M$ is the total number of agents in the traffic graph.

\subsection{Simulation Process}

During the simulation process, we use the predicted future position distribution to calculate the next position of each agent. Firstly, we sample from the distribution, and then project each sampled position onto the nearest point on the road. Finally, we smooth the projected trajectory with a linear-quadratic regulator (LQR)~\citep{aastrom2021feedback} to ensure the feasibility and smoothness of the simulated trajectory. The LQR algorithm can efficiently minimize the total commutative quadratic cost of a linear dynamic system. We consider a finite-horizon, discrete-time linear system with dynamics described by:
\begin{equation}
\left[\begin{array}{c}
    \tilde{\vp}^m_{t+1} \\
    \tilde{\vv}^m_{t+1} \\
    \end{array}\right] 
    =\left[\begin{array}{ccc}
            \mI & \mD \\
            0 & \mI\\
            \end{array}\right] 
            \left[\begin{array}{c}
                \tilde{\vp}^m_t \\
                \tilde{\vv}^m_{t} \\
                \end{array}\right] 
            + \left[\begin{array}{c}
                \mD^2 \\
                \mD
                \end{array}\right] \tilde{\va}^m_t,
\end{equation}
where $\mD$ is a diagonal matrix with the interval of each time step as diagonal entries, and $\tilde{\vp}_t^m$, $\tilde{\vv}^m_t, \tilde{\va}^m_t$ represent the LQR-planned position, velocity, and acceleration, respectively. The system is subject to a quadratic cost function:
\begin{equation}
    \mathcal{J}=\sum_{m=1}^{M}\sum_{t=1}^{T} \|\tilde{\vp}_t^m-\bar{\vp}^m_t\|^2+\eta_{\va}\|\tilde{\va}_t^m\|^2,
\end{equation}
where the projected predicted position $\bar{\vp}_{m}^{t}$ is considered as the target pose, and the hyper-parameter $\eta_{\va}$ is used to penalize high acceleration. After the LQR optimization, each agent is updated to the first position of the planned trajectory.

\section{Experiment}

\subsection{Dataset}
We use a real-world dataset called \textbf{pNEUMA}~\citep{barmpounakis2020pNEUMA} to construct a realistic urban traffic simulator. This dataset contains over half a million trajectories of various types of vehicles, collected by 10 drones in Athens over 4 days. The drones recorded traffic streams in a large area with over 100 km of lanes in the road network and around 100 busy intersections (signalized or not). However, since the dataset did not provide traffic light states, we designed an algorithm to estimate this information from the recorded trajectory data, which is described in the appendix.

The recordings were done at 5 periods during each day, spanning about 15 minutes each, with a time interval of collected data being 0.04 seconds. To enhance computation efficiency, we use a time step of 0.4 seconds. We split the dataset into a training set (recordings from the first 3 days) and a validation/test set (recordings from the last day). We do not use other popular datasets for traffic simulation, such as NGSIM~\citep{Colyar2017ngsim} or HighD~\cite{krajewski2018highd}, because they only contain simple highway scenarios without traffic lights, making it challenging to develop a generalizable urban traffic simulator.

\subsection{Metrics}

We evaluate the realism of our simulator by measuring the similarity between the simulation result and real data. During evaluation, we assume that each vehicle enters the simulator at its first recorded time and position, and is then controlled by our simulator to complete its recorded route. When an agent reaches its final recorded position, it is removed from the simulator.

Firstly, we follow prior works~\cite{song2018multi,Bhattacharyya2019SimulatingEP} and conduct a \textbf{short-term microscopic} evaluation by simulating for 20 seconds from a random time step in the test dataset. We measure the similarity between the simulated and real data using \textbf{position and velocity RMSE} metrics, which are calculated by: 
\begin{equation}
    \operatorname{RMSE}=\frac{1}{T_s}\sum_{t=1}^{T_s} \sqrt{\frac{1}{M}\sum_{m=1}^M \|\vs_t^m-\hat{\vs}_t^m\|^2},
\label{eq:rmse}
\end{equation}
where $\vs_t^m$ and $\hat{\vs}_t^m$ were the real and simulated value of the position or velocity of the agent $m$ at time step $t$, respectively. $T_s$ was the total simulated time steps, and $M$ was the total simulated agent number. We also calculate the \textbf{off-road rate}, which measures the avarage proportion of vehicles that deviate more than 1.5 meters from the road over all time steps. We does not measure the common collision rate metric because we focus on the long-term influence of the traffic model and the dataset does not provide accurate vehicle size and heading information.

In addition, we also evaluate our model's \textbf{long-term macroscopic} accuracy on five periods in the test dataset for 800 seconds from its initial recording time. To measure the long-term performance, we use two common macroscopic metrics for traffic flow data~\citep{Lighthill1955OnKW,Richards1956ShockWO,sewall2010continuum}, namely \textbf{road density and speed RMSE}, in addition to the \textbf{off-road rate}. The density of a road at a time step is calculated by dividing the number of vehicles on the road by its total lane length, assuming that all lanes have the same width. Meanwhile, the road speed is computed as the mean speed of all vehicles on the road. To quantify the similarity between the simulated and ground truth values, we still use RMSE in~\cref{eq:rmse}, where the variable $M$ becomes the total number of roads in the recorded area.

\subsection{Performance}

We conduct a comparative analysis of our method against several baselines, including one rule-based method (SUMO), two GAIL-based methods (PS-GAIL and RAIL), and two BC-based methods (BC and PS-CCIL):

\textbf{SUMO}~\citep{krajzewicz2012sumo}: we use the IDM model~\cite{treiber2000idm} as the car-following model and mobil~\citep{kesting2007general} as the lane-changing model. We tune the IDM's parameters by minimizing the MSE between the IDM calculated acceleration and real acceleration using an Adam optimizer~\cite{Kingma2014AdamAM}.

\textbf{PS-GAIL}~\citep{song2018multi}: we learn our model based on GAIL and let all vehicles share the same policy parameter and critic parameter with the PPO~\citep{Schulman2017ProximalPO} as the reinforcement learning algorithm.

\textbf{RAIL}~\citep{Bhattacharyya2019SimulatingEP}: we use the same learning process as \textbf{PS-GAIL} but with an additional off-road penalty.

\textbf{BC}~\citep{Michie1990BC}: we learn our model structure directly by BC without removing the context history or perturbing the historical trajectory.

\textbf{Parameter-Sharing CCIL (PS-CCIL)}: we directly extend the CCIL method~\citep{Guo2023CCIL} by making all vehicle sharing the same policy parameter without removing the context history.

We train and evaluate each model three times to obtain the mean and standard deviation (std) of various metrics. We evaluate both short-term and long-term performance, as shown in Tables~\ref{tab:short} and~\ref{tab:long}, respectively. Our method achieves better results than all baselines in terms of position and velocity RMSE, road density and speed RMSE, with minor off-road rate.

\begin{table*}[t]
\caption{Comparison with baselines on microscopic metrics for 20 seconds}
\begin{center}
\begin{tabular}{l|cccc}
\textbf{Model} & \textbf{Position RMSE(m)} & \textbf{Velocity RMSE(m/s)} & \textbf{Off-road(\%)}  \\ \toprule 
SUMO~\citep{krajzewicz2012sumo} & 41.25 & 7.00 & \textbf{0}  \\
PS-GAIL~\citep{song2018multi} & 61.65\textpm2.56 & 6.67\textpm0.32 & 1.72\textpm0.13 \\
RAIL~\citep{Bhattacharyya2019SimulatingEP} & 55.78\textpm2.47  & 5.93\textpm0.19  &  0.59\textpm0.04 \\
BC~\citep{Michie1990BC} & 39.95\textpm 1.53  &  6.60\textpm0.23 & 31.80\textpm 2.12 \\
PS-CCIL & 20.45\textpm0.51 & 3.80\textpm0.05 & 0.45\textpm0.02\\
\textbf{HMMIL (ours)} & \textbf{20.10\textpm 0.65} & \textbf{3.72\textpm0.08} & 0.48\textpm 0.01\\
\bottomrule
\end{tabular}
\end{center}
\vspace{-0.4cm}
\label{tab:short}
\end{table*}

\begin{table*}[t]
\caption{Comparison with baselines and ablated models on macroscopic metrics for 800 seconds}
\vspace{-0.3cm}
\begin{center}
\begin{tabular}{l|ccc}
\textbf{Model} & \textbf{Road Density RMSE(veh/km)} & \textbf{Road Speed RMSE(m/s)} & \textbf{Off-road(\%)}  \\ \toprule 
SUMO~\citep{krajzewicz2012sumo} & 52.70 & 5.52 & \textbf{0}  \\
PS-GAIL~\cite{song2018multi} & 54.06\textpm1.23 & 4.03\textpm0.05 & 13.24\textpm3.20 \\
RAIL~\cite{Bhattacharyya2019SimulatingEP} & 54.45\textpm1.89 & 3.89\textpm0.11 & 2.92\textpm0.38 \\
BC~\citep{Michie1990BC} & 61.51\textpm 1.53  &  5.38\textpm0.21 & 42.15\textpm 5.25 \\
PS-CCIL & 62.11\textpm0.43 & 3.88\textpm0.08 & 0.53\textpm0.03 \\
\textbf{HMMIL (ours)} & 48.71\textpm0.18 &  \textbf{3.55 \textpm 0.13} & 0.51\textpm 0.02 \\
\bottomrule
\end{tabular}
\end{center}
\vspace{-0.7cm}
\label{tab:long}
\end{table*}







\subsection{Qualitative Result}
\begin{figure*}[t]
	\centering
	\includegraphics[width=\linewidth]{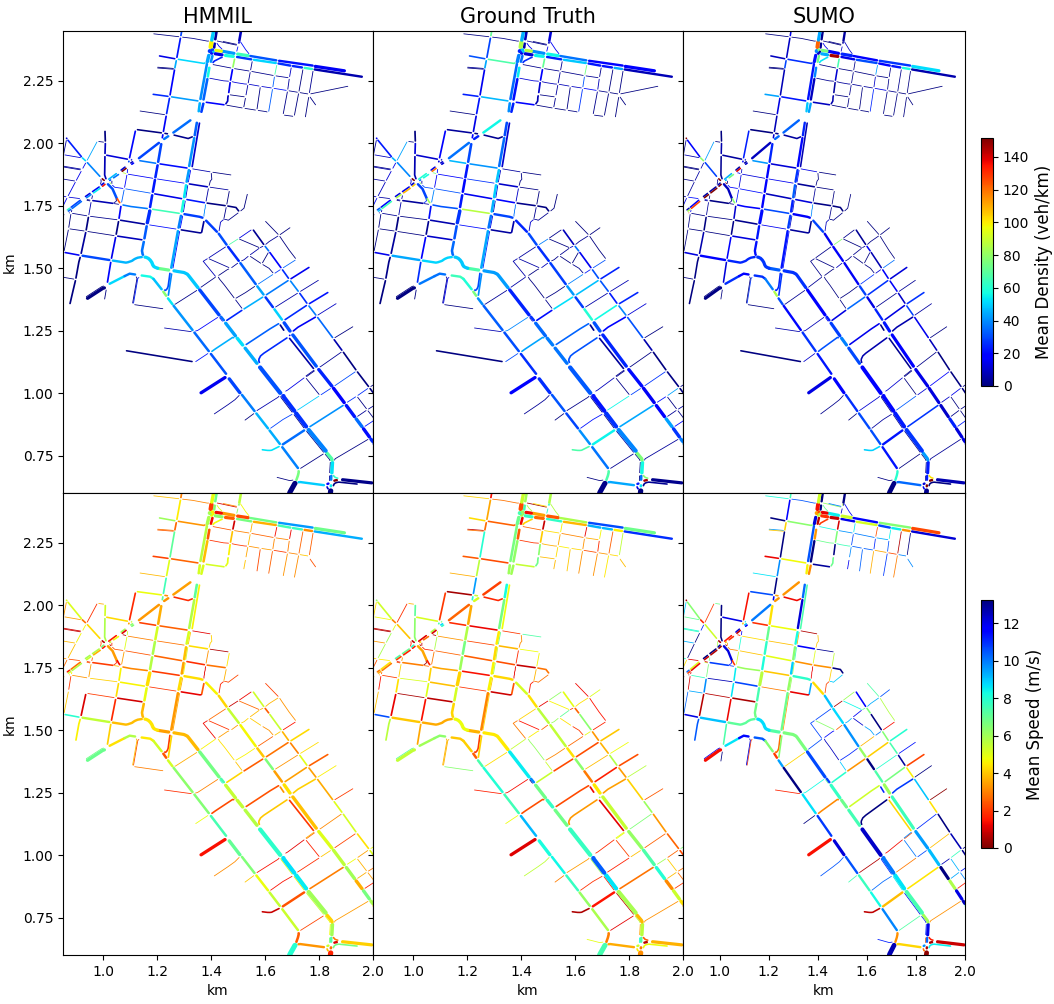}
	\vspace{-0.7cm}
	\caption{Mean density and speed on each road over all time steps in the long-term evaluation.}
	\label{fig:densityspeed}
\end{figure*}

In~\cref{fig:densityspeed}, we present the mean road density and speed for real-world data, SUMO simulation, and our proposed method over all time steps. The figure shows that our proposed method accurately reproduces long-term macroscopic traffic patterns, outperforming the SUMO simulator.

\section{Limitation}

There are two main limitations of our simulator. Firstly, even though removing all history information is a simple way to create a long-term simulation, some history information may be important for microscopic decision-making. To create a more realistic human driving model with history information input, a new method that can address the \textit{covariate shift} problem in the history and context needs to be developed, as existing GAIL-based multi-agent imitation learning methods perform poorly in practice. Secondly, although the \textbf{pNEUMA} dataset records a large area with thousands of vehicles, it does not provide accurate vehicle shape, heading, high-definition map, and traffic light information. Therefore, our learned model's microscopic performance is limited by the data accuracy. Future work could focus on improving the data accuracy or exploring alternative data sources to address the limitation.

\section{Conclusion}

In conclusion, we propose a history-masked multi-agent imitation learning method for realistic long-term microscopic traffic simulation. Our method addresses the \textit{covariate shift} issue in multi-agent imitation learning, allowing us to generate stable long-term traffic simulations that are essential for transportation planning and developing more traffic-aware autonomous driving policies. We apply our approach specifically to the urban traffic simulation problem and achieve better short-term microscopic and long-term macroscopic similarity to real-world data than state-of-the-art baselines.



\bibliography{paper}  

\end{document}